\documentclass[runningheads]{llncs}
\usepackage[T1]{fontenc}
\usepackage{comment}
\usepackage{amsmath}
\usepackage{amssymb}
\usepackage{graphicx}
\usepackage{xcolor}
\usepackage{colortbl}
\usepackage{multirow}
\usepackage{booktabs}
\usepackage{enumitem}
\usepackage{algorithm}
\usepackage{algorithmic}
\usepackage{subcaption}
\usepackage{caption}
\usepackage[colorlinks=true, allcolors=blue]{hyperref}

\definecolor{LightPink}{rgb}{1.0, 0.8, 0.8} 
\definecolor{Peach}{rgb}{1.0, 0.9, 0.8} 
\begin{document}
\title{ Metric Embedding Initialization-Based Differentially Private and Explainable \\
Graph Clustering }
\titlerunning{Private and Explainable Graph Clustering}
%
%

\author{
Haochen You\inst{1}\orcidID{0009-0008-9178-2912}\thanks{Corresponding author.} \and
Baojing Liu\inst{2}\orcidID{0009-0007-1444-7267}
}

\authorrunning{H. You and B. Liu}

\institute{
Graduate School of Arts and Sciences\\
Columbia University, New York, USA\\
\email{hy2854@columbia.edu}
\and
School of Artificial Intelligence\\
Hebei Institute of Communications, Shijiazhuang, PR China\\
\email{liubj@hebic.edu.cn}
}

%
%
%
\maketitle              
\begin{abstract}

Graph clustering under the framework of differential privacy, which aims to process graph-structured data while protecting individual privacy, has been receiving increasing attention. Despite significant achievements in current research, challenges such as high noise, low efficiency and poor interpretability continue to severely constrain the development of this field. In this paper, we construct a differentially private and interpretable graph clustering approach based on metric embedding initialization. Specifically, we construct an SDP optimization, extract the key set and provide a well-initialized clustering configuration using an HST-based initialization method. Subsequently, we apply an established k-median clustering strategy to derive the cluster results and offer comparative explanations for the query set through differences from the cluster centers. Extensive experiments on public datasets demonstrate that our proposed framework outperforms existing methods in various clustering metrics while strictly ensuring privacy.


\end{abstract}

\section{Introduction}

Differential Privacy (DP) is a mathematical framework for protecting data privacy, widely applied in fields such as statistical analysis and machine learning \cite{dwork2006differential}. Its core goal is to safeguard individual privacy while providing useful information, ensuring that even with external knowledge, an attacker cannot infer sensitive details about a single individual from query results. DP achieves this by introducing noise through mechanisms such as the Laplace, Gaussian, and exponential mechanisms, and has been widely adopted in real-world scenarios due to strong theoretical guarantees. However, challenges remain in addressing the trade-off between noise-induced utility loss and analysis accuracy, efficiently handling non-standard data, and managing dynamic data like time series \cite{dwork2008differential}.



We focus on graph clustering under the framework of differential privacy, an important problem at the intersection of graph theory and data mining \cite{schaeffer2007graph}. This framework aims to group the nodes in a graph such that nodes within the same group are closely connected, while nodes in different groups are sparsely connected. Graph structures have significant natural advantages in representing data from various modern societal scenarios, but their high dimensionality, sparsity, and scalability pose challenges for designing corresponding algorithms~\cite{dwork2006differential}.

Spectral methods are one of the key approaches to addressing the above issues, achieving great success in graph clustering problems \cite{bernardi1997spectral}. However, their application under the differential privacy framework has yet to be deeply explored. Meanwhile, in existing clustering algorithms, particularly those based on iterative optimization, the choice of initial centers is highly sensitive \cite{ge2023optimally,he2024differentially}. For graph clustering under the framework of differential privacy, there are additional unique challenges, such as high-dimensional data distribution, non-uniformity, high noise, and application-specific requirements (private dataset), which impose even higher demands on the initialization of centers \cite{wang2021fast}.

Our main contributions are given:
\begin{enumerate}[label=(\roman*)] 
    \item We propose a novel graph clustering framework under differential privacy, integrating spectral methods and semidefinite programming.
    \item We combined HST to provide an initialization scheme for the  k-median methods used in graph clustering and presented detailed algorithmic steps.
    \item We conducted extensive numerical experiments on public datasets, demonstrating the effectiveness and indispensability of each module of our model.
\end{enumerate}

\section{Private and Explainable Graph Clustering}

Let $G = (V, E)$ be the input graph. Let $u, v \in V$ denote two nodes in the graph. We denote $m = |E|$ as the number of edges and $d_G(u)$ as the degree of node $u$ in $G$. The parameter $\lambda > 0$ is a user-defined coefficient controlling the trade-off between structural fidelity and cluster separation. The variable $b$ quantifies the balance between cluster volumes. We use SDP to extract the graph clustering structure. Let $\overline{\boldsymbol{u}}, \overline{\boldsymbol{v}} \in \mathbb{R}^d$ denote the vector embeddings of nodes $u$ and $v$. The optimization expression is as follows:

\begin{equation}
\begin{aligned}
\min & \sum_{(u, v) \in E}\|\overline{\boldsymbol{u}}-\overline{\boldsymbol{v}}\|_2^2
+ \frac{2 \sum_{u, v \in V}\langle\overline{\boldsymbol{u}}, \overline{\boldsymbol{v}}\rangle^2 \mathrm{~d}_G(u) \mathrm{d}_G(v)}{\lambda m} \\
\text {s.t.} & \sum_{u, v \in V}\left(\|\overline{\boldsymbol{u}}-\overline{\boldsymbol{v}}\|_2^2 \mathrm{~d}_G(u) \mathrm{d}_G(v)\right) \geq 2 b m^2, \\
& \langle\overline{\boldsymbol{u}}, \overline{\boldsymbol{v}}\rangle \geq 0, \text { for all } u, v \in V; \|\overline{\boldsymbol{u}}\|_2^2 = 1, \text { for all } u \in V,
\end{aligned}
\label{eq:optimization_problem}
\end{equation}
where $b=\frac{1}{m^2} \sum_{i, j \in[k], i \neq j} \operatorname{vol}_G\left(C_i\right) \operatorname{vol}_G\left(C_j\right)=1-\frac{1}{2 m^2} \sum_{i \in[k]} \operatorname{vol}_G\left(C_i\right)^2$.

To facilitate efficient optimization, we reformulate the vector-based objective into matrix form. Specifically, we define the unnormalized graph Laplacian as $L_G$, and the Gram matrix $X \in \mathbb{R}^{n \times n}$ with $X_{i,j} = \frac{1}{n} \bar{v}_i \cdot \bar{v}_j$, where $\bar{v}_i$ is the embedding of node $i$. Under this formulation, node similarities are captured via inner products, and the trace term $\langle L_G, X \rangle$ corresponds to the spectral cut cost \cite{chen2014improved}. The regularization term encourages balanced clustering through degree-normalized representations. The resulting optimization problem is:
\begin{equation}
    \min\limits_{\boldsymbol{X} \in \mathcal{D}}  \left\langle\boldsymbol{L}_{\boldsymbol{G}}, \boldsymbol{X}\right\rangle+\frac{n}{\lambda m}\left\|\boldsymbol{D}_{\boldsymbol{G}}^{1 / 2} \boldsymbol{X} \boldsymbol{D}_{\boldsymbol{G}}^{1 / 2}\right\|_{\mathrm{F}}^2 ,
\end{equation}
where the feasible region $\mathcal{D}$ is defined as:
\begin{equation*}
    \mathcal{D} := \left\{ \left\langle\boldsymbol{D}_{\boldsymbol{G}} \boldsymbol{L}_{V} \boldsymbol{D}_{\boldsymbol{G}}, \boldsymbol{X}\right\rangle \geq \frac{b m^2}{n}, \boldsymbol{X} \succeq 0, \boldsymbol{X} \geq 0, \boldsymbol{X}_{i i}=\frac{1}{n}, \forall i \right\}.
\end{equation*}

We add Gaussian noise $W$ to the matrix and use spectral decomposition to provide a graph embedding for each vertex. The addition of Gaussian noise $W$ to the matrix $\boldsymbol{X}_1$ ensures $(\varepsilon, \delta)$-differential privacy via the Gaussian mechanism. Since the sensitivity of the spectral embedding process is bounded under the Frobenius norm, the noise scale is calibrated accordingly~\cite{dwork2008differential}.

\begin{algorithm}[htbp]
    \caption{Privacy-Preserving Spectral Embedding.}
    \label{Privacy-Preserving Spectral Embedding}
    \begin{algorithmic}
        \STATE \textbf{Input}: Global rep. $X_1$, privacy parameters $\varepsilon, \delta$, target number of clusters $k$
        \STATE $\boldsymbol{W} \leftarrow \mathcal{N}\left(0,24(\lambda+3) m \cdot \frac{\ln (2 / \delta)}{\epsilon^2}\right)^{n \times n}$
        \STATE $\boldsymbol{X}_{\mathbf{2}} \leftarrow n \boldsymbol{D}_{\boldsymbol{G}}^{1 / 2} \boldsymbol{X}_{\mathbf{1}} \boldsymbol{D}_{\boldsymbol{G}}^{1 / 2}+\boldsymbol{W}$
        \STATE $\boldsymbol{f}_1, \dots, \boldsymbol{f}_k \leftarrow \text{eigenvectors of } \boldsymbol{X}_2 \text{ for its } k \text{ smallest eigenvalues}$
        \STATE $F (V(G) \rightarrow \mathbb{R}^k) \leftarrow \mathrm{d}_G(u)^{-1 / 2}\left(\boldsymbol{f}_{\mathbf{1}}(u), \boldsymbol{f}_{\mathbf{2}}(u), \ldots, \boldsymbol{f}_{\boldsymbol{k}}(u)\right)^{\top}$
        \STATE \textbf{Output}: Embedding representation of vertices $F$
    \end{algorithmic}
\end{algorithm}

We utilize the DP exponential mechanism to construct a private ranking set~\cite{ghazi2020differentially}, scale the low-dimensional clustering cost to estimate the original high-dimensional cost, and recover the centroid structure. The steps are presented in Algorithm~\ref{Critical Set Generation}. The HST-based initialization introduces Laplace noise to node counts during subtree scoring. As each node query has bounded sensitivity and the Laplace mechanism is applied independently at each level of the tree, the algorithm achieves $(\varepsilon, 0)$-differential privacy.

To implement this initialization strategy, we adopt a HST structure to represent the recursive partitioning of the data space. Specifically, the data point set is recursively divided, with each partition forming smaller subsets. Each node at level $i$ represents a partition of diameter $\Delta / 2^{L-i}$, where $\Delta$ is the initial diameter and $L$ is the tree depth. The top-level node corresponds to the full dataset, its children represent first-level partitions, and the process continues until each leaf represents a single data point. The resulting HST structure enables both efficient initialization and privacy-preserving subtree scoring~\cite{fan2024k}.

The result of this construction is a well-separated HST, where each node represents a partition, and the leaf nodes represent the specific data points \cite{HaochenYou}. The detailed steps are presented in Algorithm \ref{Subtree and leaf search}.

\begin{algorithm}[htbp]
    \caption{Critical Set Generation.}
    \label{Critical Set Generation}
    \begin{algorithmic}
        \STATE \textbf{Input}: Data points $\{ x_i \}_n$, primitive dimension $d$, dimension after reduction $d^{\prime}$, target \# clusters $k$, privacy parameter $\epsilon$, error parameters $\lambda, \alpha, \beta$, distance parameter $p$
        \STATE $\zeta \leftarrow 0.01\left(\frac{\alpha}{10 \lambda_{p, \alpha / 2}}\right)^{p / 2}, \Lambda \leftarrow \sqrt{\frac{0.01 d}{\ln (n / \beta) d^{\prime}}}$
        \STATE \textbf{for} $i \in\{1, . ., n\}$ \textbf{do}
        \STATE \hspace{2em} $\tilde{x}_i \leftarrow \Pi_{\mathcal{S}}\left(x_i\right)$
        \STATE \hspace{2em} \textbf{if} $\left\|\tilde{x}_i\right\| \leq 1 / \Lambda$ \textbf{then}
        \STATE \hspace{4em}  $x_i^{\prime}=\Lambda \tilde{x}_i$
        \STATE \hspace{2em} \textbf{else}
        \STATE \hspace{4em}  $x_i^{\prime}=0$
        \STATE \textbf{end for}
        \STATE C $\leftarrow PCS ^{\epsilon / 2}\left(x_1^{\prime}, \ldots, x_n^{\prime}, \zeta\right)$ \cite{ghazi2020differentially}, $cost\left(S_{\epsilon} \right) \leftarrow \left(\frac{\ln (n / \beta)}{0.01}\right)^{p / 2} NPA (C, k)$ \cite{makarychev2019performance}
        \STATE \textbf{Output}: Critical set $ C $, $cost(S_{\epsilon})$
    \end{algorithmic}
\end{algorithm}

We construct $k$ initial cluster centers from the HST by selecting $k$ high-score nodes as subtree roots, using $score(v) = N_v \cdot 2^{h_v}$, where $N_v$ is the number of data points and $h_v$ is the level of node $v$. To ensure diversity, selected nodes have no ancestor-descendant relations. If candidates are insufficient, the process repeats. Within each subtree, a center is chosen by greedily descending to the leaf with the highest score. Under differential privacy, Laplace noise is added to node counts before selection. The full procedure is shown in Algorithm~\ref{Finding Initial Cluster Centers}.

\begin{algorithm}[htbp]
    \caption{Subtree and leaf search.}
    \label{Subtree and leaf search}
    \begin{algorithmic}
        \STATE \textbf{Input}: Hierarchical well-separated tree $T$, depth of tree $L$, privacy-protected data point set $C'$, target number of clusters $k$
        \STATE $C_0 \leftarrow \emptyset, C_1 \leftarrow \emptyset$
        \STATE \textbf{for} each node $v$ in $T$ \textbf{do}
        \STATE \hspace{2em} $N_v \leftarrow|C' \cap T(v)| + Lap(2^{L-h_v}/\epsilon)$, $score(v) \leftarrow N(v) \cdot 2^{h_v}$
        \STATE \textbf{end for}
        \STATE \textbf{while} $\left|C_1\right|<k$ \textbf{do}
        \STATE \hspace{2em} Add top $\left(k-\left|C_1\right|\right)$ nodes with highest score to $C_1$
        \STATE \hspace{2em} \textbf{for} each $v \in C_1$ \textbf{do}
        \STATE \hspace{2.5em} $C_1=C_1 \backslash\{v\}$, if $\exists \: v^{\prime} \in C_1$ such that $v^{\prime}$ is a descendant of $v$
        \STATE \hspace{2em} \textbf{end for}
        \STATE \textbf{for} each node $v$ in $C_1$ \textbf{do}
        \STATE \hspace{2em} \textbf{while} $v$ is not a leaf node \textbf{do}
        \STATE \hspace{4em} $v \leftarrow \arg _w \max \left\{N_w, w \in \operatorname{ch}(v)\right\}$, $c h(v)$ denotes the children nodes of $v$
        \STATE \hspace{2em} Add $v$ to $C_0$
        \STATE \textbf{end for}
        \STATE \textbf{Output}: Center set of subtree $C_0$
    \end{algorithmic}
\end{algorithm}

\begin{algorithm}[htbp]
    \caption{Finding Initial Cluster Centers.}
    \label{Finding Initial Cluster Centers}
    \begin{algorithmic}
        \STATE \textbf{Input}: Data point set $C$
        \STATE $\triangle \leftarrow \max\limits_{c_1, c_2 \in C} \| c_1 - c_2 \|, L \leftarrow \ln \triangle$
        \STATE Randomly pick a point in $C$ as the root node
        \STATE \textbf{for} each $c \in C$ \textbf{do}
        \STATE \hspace{2em} Set $V_c=[c]$
        \STATE \hspace{2em} \textbf{for} each $c' \in C$ \textbf{do}
        \STATE \hspace{4em} Add $c' \in C$ to $V_c$ if $d(c, c') \leq \triangle/2$ and $c' \notin \bigcup_{v \neq c} C_{v}$
        \STATE \hspace{2em} \textbf{end for}
        \STATE \textbf{end for}
        \STATE Set the non-empty clusters $V_c$ as the children nodes of $T$
        \STATE \textbf{for} each non-empty cluster $V_c$ \textbf{do}
        \STATE \hspace{2em} Run 2-HST $\left(V_c, L-1\right)$ to extend the tree $T$
        \STATE \hspace{2em} stop until $L$ levels or reaching a leaf node
        \STATE \textbf{end for}
        \STATE $C_0 \leftarrow$ \textbf{Algorithm} \ref{Subtree and leaf search} ($T, L$)
        \STATE \textbf{Output}: Private initial center set $C_0 \subseteq C$
    \end{algorithmic}
\end{algorithm}

After completing the aforementioned key steps, we finally introduce the explainability module. Using the original clustering cost from Algorithm \ref{Critical Set Generation}, we incorporate the fixed-centroid clustering algorithm to restore the clustering structure and cost in the high-dimensional space. The clustering cost is then calculated as a contrastive explanation for query users based on the fixed-centroid results \cite{nguyen2024contrastive}. The final integrated algorithm steps are shown in Algorithm \ref{pgc}.

\begin{algorithm}[htbp]
    \caption{Private and Explainable Graph Clustering.}
    \label{pgc}
    \begin{algorithmic}
        \STATE \textbf{Input}: $G=(V, E)$, target \# clusters $k$, query user set $V_s$, error parameter $\beta$
        \STATE $\boldsymbol{X}_{\mathbf{1}} \leftarrow \arg \min _{\boldsymbol{X} \in \mathcal{D}}\left\langle\boldsymbol{L}_{\boldsymbol{G}}, \boldsymbol{X}\right\rangle+\frac{n}{\lambda m}\left\|\boldsymbol{D}_{\boldsymbol{G}}^{1 / 2} \boldsymbol{X} \boldsymbol{D}_{\boldsymbol{G}}^{1 / 2}\right\|_{\mathrm{F}}^2$
        \STATE $F \leftarrow$ \textbf{Algorithm} \ref{Privacy-Preserving Spectral Embedding} $(X_1)$
        \STATE Critical Set $C$ , $cost(S_{\epsilon})$ $\leftarrow$ \textbf{Algorithm} \ref{Critical Set Generation} $(F(u), u \in V)$
        \STATE Private initial center set $C_0 \subseteq C \leftarrow$ \textbf{Algorithm} \ref{Finding Initial Cluster Centers} $(C)$
        \STATE $\widehat{C}_1, \ldots, \widehat{C}_k \leftarrow$ KMedian $(F(c), c \in C \phantom{b} | \phantom{b} \text{Initial center set} \phantom{b} C_0)$
        \STATE $ Exp(i) \leftarrow | cost(S_{\epsilon}) - \left(\frac{\ln (n / \beta)}{0.01}\right)^{\frac{p}{2}} FixedCenter(C,k,x_i) |$ \cite{charikar1999constant,kanungo2002local}
        \STATE \textbf{Output}:  $k$-partition $\left\{\widehat{C}_i\right\}$ , comparative explanation collection $\{ Exp(i) | x_i \in V_s \}$
    \end{algorithmic}
\end{algorithm}

\section{Experiments}

First, we aim to verify the effectiveness of the metric embedding-based clustering center initialization module (\textbf{MEI}), specifically Algorithm \ref{Finding Initial Cluster Centers}, through comparative experiments. The baseline models we selected include: \textbf{DPFN} \cite{romijnders2024protect}, \textbf{BR-DP} \cite{jiang2024budget}, \textbf{PP-DOAGT} \cite{huang2024differential}, \textbf{QFL-DP} \cite{rofougaran2024federated}, \textbf{DNN-SDP} \cite{kumar2024differential}. Due to space limitations, please refer to the arXiv version for detailed experimental setup.

\begin{figure}[htbp]
    \centering
    \begin{subfigure}[t]{0.45\linewidth}
        \centering
        \includegraphics[width=0.8\linewidth]{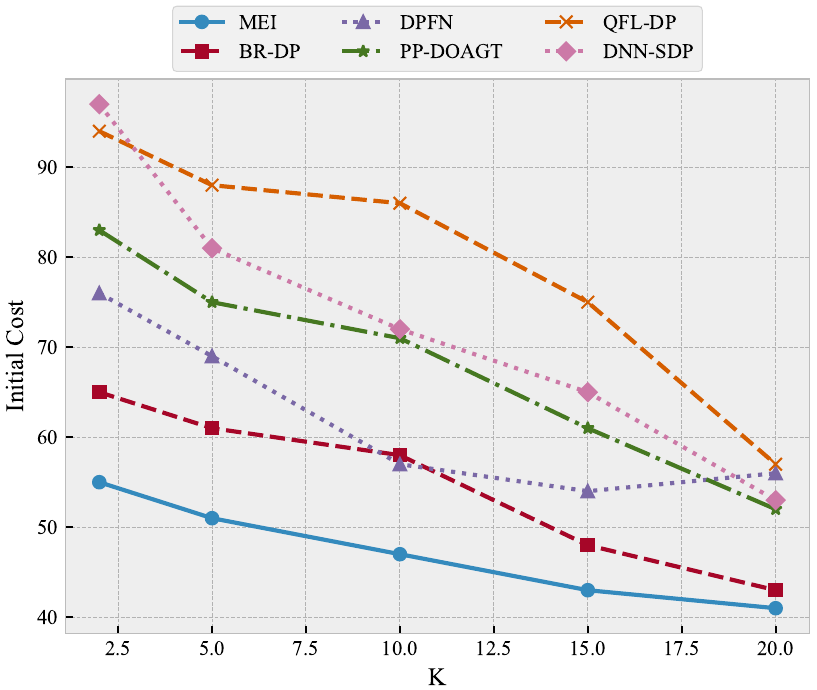}
        \caption{The initial cost.}
        \label{Initial Cost}
    \end{subfigure}
    \hspace{0\linewidth}
    \begin{subfigure}[t]{0.45\linewidth}
        \centering
        \includegraphics[width=0.8\linewidth]{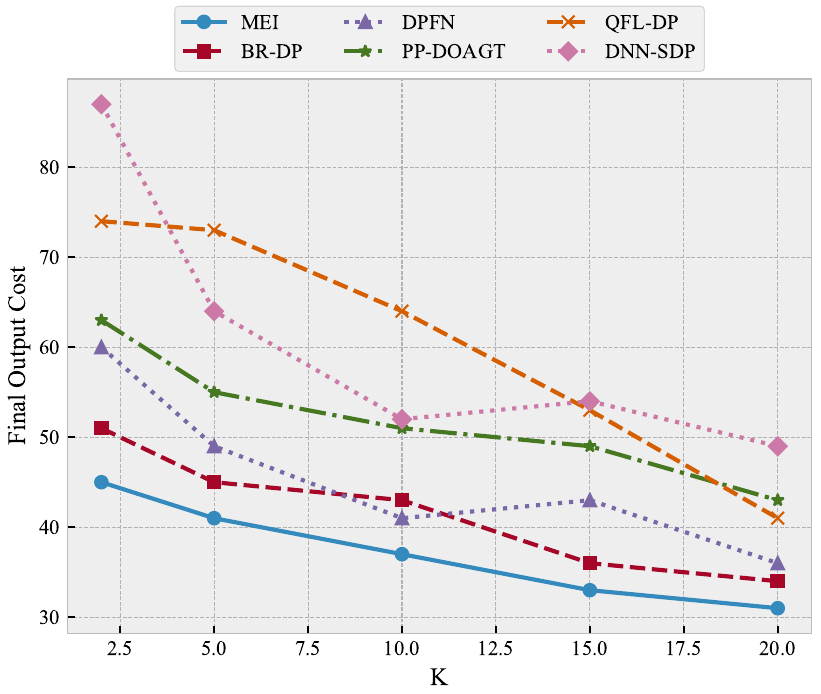}
        \caption{The final output cost.}
        \label{Final Output Cost}
    \end{subfigure}
    \caption{The cost of graph metric experiment.}
    \label{fig:two_images}
\end{figure}

As shown in \ref{Initial Cost} and \ref{Final Output Cost}, it can be observed that both the initial cost values at the start of the algorithm and the terminal values at the end of the iterations are significantly lower for our algorithm compared to other similar algorithms. This demonstrates its superior ability in identifying good initial clustering centers and providing favorable preconditions for the subsequent clustering process.

Next, we conduct experiments on publicly available datasets. The selected datasets include \textbf{USPS} \cite{le1990handwritten}, \textbf{Reuters} \cite{lewis2004rcv1}, \textbf{DBLP} \footnote{\url{https://dblp.uni-trier.de/}}, \textbf{ACM} \footnote{\url{https://dl.acm.org/}}, \textbf{CiteSeer} \footnote{\url{https://paperswithcode.com/dataset/citeseer}}, and \textbf{HHAR} \cite{stisen2015smart}. Comparable baselines were selected, including $(\epsilon,\delta)$-\textbf{DP} \cite{he2024differentially}, \textbf{PE} \cite{fan2024k}, \textbf{NpGC} \cite{yu2024non}, \textbf{NMFGC} \cite{jannesari2024novel}, \textbf{SCGC} \cite{kulatilleke2025scgc}. From the data in Table \ref{Overall Performance}, it can be seen that our proposed model achieves significant improvements over existing models in most metrics while ensuring privacy protection.


\begin{table*}[t]
\centering
\caption{Overall Performance on NMI, Purity, ACC, ARI, and F1.}
\label{Overall Performance}
\resizebox{\textwidth}{!}{
\begin{tabular}{l|ccccc|ccccc}
\toprule
Method & \multicolumn{5}{c}{\textbf{USPS}} &  \multicolumn{5}{c}{\textbf{Reuters}} \\
\midrule
$(\epsilon,\delta)$-\textbf{DP} & 0.801 & 0.638 & 0.830 & 0.796 & 0.805 & \cellcolor{Peach}0.557 & \cellcolor{Peach}0.791 & 0.714 & 0.578 & 0.651 \\
\textbf{PE} & \cellcolor{Peach}0.831 & 0.651 & 0.806 & \cellcolor{Peach}0.819 & 0.782 & 0.553 & 0.766 & \cellcolor{Peach}0.726 & 0.505 & 0.629 \\
\textbf{NpGC} & 0.829 & \cellcolor{Peach}0.665 & 0.815 & 0.802 & 0.741 & 0.538 & 0.782 & 0.698 & 0.513 & \cellcolor{Peach}0.673 \\
\textbf{NMFGC} & 0.783 & 0.598& \cellcolor{Peach}0.835 & 0.776 & \cellcolor{Peach}0.811 & 0.496 & 0.732 & 0.681 & 0.577 & 0.648 \\
\textbf{SCGC} & 0.814 & 0.647 & 0.825 & 0.799 & 0.764 & 0.510 & 0.773 & 0.709 & \cellcolor{Peach}0.582 & 0.635 \\
\textbf{Our Model} & \cellcolor{LightPink} \textbf{0.859} & \cellcolor{LightPink}\textbf{0.691} & \cellcolor{LightPink}\textbf{0.847} & \cellcolor{LightPink}\textbf{0.805} & \cellcolor{LightPink}\textbf{0.832} & \cellcolor{LightPink}\textbf{0.565} & \cellcolor{LightPink}\textbf{0.807} & \cellcolor{LightPink}\textbf{0.743} & \cellcolor{LightPink}\textbf{0.594} & \cellcolor{LightPink}\textbf{0.680} \\
\midrule\midrule
Method & \multicolumn{5}{c}{\textbf{DBLP}} &  \multicolumn{5}{c}{\textbf{ACM}} \\
\midrule
$(\epsilon,\delta)$-\textbf{DP} & \cellcolor{Peach}0.483 & 0.462 & 0.712 & \cellcolor{Peach}0.509 & 0.740 & 0.718 & \cellcolor{Peach}0.651 & \cellcolor{Peach}0.933 & 0.749 & 0.901 \\
\textbf{PE} & 0.479 & 0.473 & 0.758 & 0.496 & 0.732 & 0.722 & 0.610 & 0.905 & 0.748 & 0.881 \\
\textbf{NpGC} & 0.466 & 0.415 & 0.735 & 0.478 & 0.694 & 0.709 & 0.617 & 0.931 & 0.754 & \cellcolor{Peach}0.908 \\
\textbf{NMFGC} & 0.457 & \cellcolor{Peach}0.478 & \cellcolor{Peach}0.764 & 0.469 & 0.750 & 0.688 & 0.647 & 0.862 & \cellcolor{Peach}0.771 & 0.870 \\
\textbf{SCGC} & 0.476 & 0.458 & 0.741 & 0.475 & \cellcolor{Peach}0.760 & \cellcolor{Peach}0.727 & 0.650 & 0.919 & 0.764 & 0.889 \\
\textbf{Our Model} & \cellcolor{LightPink} \textbf{0.476} & \cellcolor{LightPink}\textbf{0.493} & \cellcolor{LightPink}\textbf{0.784} & \cellcolor{LightPink}\textbf{0.518} & \cellcolor{LightPink}\textbf{0.762} & \cellcolor{LightPink}\textbf{0.733} & \cellcolor{LightPink}\textbf{0.679} & \cellcolor{LightPink}\textbf{0.956} & \cellcolor{LightPink}\textbf{0.804} & \cellcolor{LightPink}\textbf{0.917} \\
\midrule\midrule
Method & \multicolumn{5}{c}{\textbf{CiteSeer}} &  \multicolumn{5}{c}{\textbf{HHAR}} \\
\midrule
$(\epsilon,\delta)$-\textbf{DP} & 0.437 & 0.554 & 0.716 & 0.488 & \cellcolor{Peach}0.647 & 0.835 & 0.703 & 0.870 & 0.812 & \cellcolor{Peach}0.902 \\
\textbf{PE} & 0.418 & 0.537 & 0.723 & 0.499 & 0.630 & \cellcolor{Peach}0.844 & 0.666 & 0.805 & 0.824 & 0.873 \\
\textbf{NpGC} & 0.426 & 0.517 & 0.708 & \cellcolor{Peach}0.517 & 0.637 & 0.840 & \cellcolor{Peach}0.705 & \cellcolor{Peach}0.880 & 0.859 & 0.891 \\
\textbf{NMFGC} & 0.469 & 0.538 & 0.719 & 0.502 & 0.646 & 0.811 & 0.647 & 0.847 & 0.845 & 0.862 \\
\textbf{SCGC} & \cellcolor{Peach}0.441 & \cellcolor{Peach}0.561 & \cellcolor{Peach}0.732 & 0.489 & 0.639 & 0.0259 & 0.828 & 0.867 & \cellcolor{Peach}0.863 & 0.884 \\
\textbf{Our Model} & \cellcolor{LightPink} \textbf{0.471} & \cellcolor{LightPink}\textbf{0.584} & \cellcolor{LightPink}\textbf{0.749} & \cellcolor{LightPink}\textbf{0.528} & \cellcolor{LightPink}\textbf{0.657} & \cellcolor{LightPink}\textbf{0.853} & \cellcolor{LightPink}\textbf{0.726} & \cellcolor{LightPink}\textbf{0.895} & \cellcolor{LightPink}\textbf{0.879} & \cellcolor{LightPink}\textbf{0.904} \\
\bottomrule
\end{tabular}
}
\end{table*}

\section{Conclusion}

We construct a differentially private and interpretable graph clustering framework based on metric embedding initialization and key set construction. Experiments on public datasets demonstrate the effectiveness and indispensability of each component of our model. Our research significantly addressing many of the current challenges and laying a solid foundation for broader applications.




%
%
%
\bibliographystyle{splncs04}
\bibliography{mybibliography}

\end{document}